\documentclass[conference]{IEEEtran}
\IEEEoverridecommandlockouts
\usepackage[utf8]{inputenc}
\usepackage[T1]{fontenc}
\usepackage{url}			 
\usepackage{cite}            

\usepackage{hyperref}
\hypersetup{ 
    colorlinks=true, 
    linkcolor=black, 
    citecolor=black, 
    urlcolor=black, 
    pdftitle={Your Title}, 
    pdfauthor={Your Name}, 
    bookmarksnumbered=true,
    bookmarksopen=true,
    breaklinks=true,
}
\usepackage[cmex10]{amsmath}  

\usepackage{mleftright}       

\usepackage{cgitIEEE}  
\mleftright

\usepackage{graphicx}         
\usepackage{booktabs}         

\usepackage[caption=false,font=footnotesize]{subfig}

\usepackage[dvipsnames]{xcolor}

\usepackage{algorithm}
\usepackage{algorithmic}

\IEEEoverridecommandlockouts

\begin{document}

\title{A Dual Optimization View to Empirical Risk Minimization with $f$-Divergence Regularization
\thanks{This work is supported in part by a University of Sheffield ACSE PGR scholarship; University of Sheffield PGR Publication Scholarship scheme; the European Commission through the H2020-MSCA-RISE-2019 project 872172; the French National Agency for Research (ANR) through the Project ANR-21-CE25-0013 and the project ANR-22-PEFT-0010 of the France 2030 program PEPR R\'eseaux du Futur; and the Agence de l'innovation de d\'efense (AID) through the project UK-FR 2024352.}
}




\author{%
 \IEEEauthorblockN{Francisco Daunas\IEEEauthorrefmark{1}\IEEEauthorrefmark{2},
                    I{\~n}aki Esnaola \IEEEauthorrefmark{1}\IEEEauthorrefmark{3},
and                    Samir M. Perlaza\IEEEauthorrefmark{2}\IEEEauthorrefmark{3}\IEEEauthorrefmark{5},
}
   \IEEEauthorblockA{\IEEEauthorrefmark{1}%
                     Sch. of Electrical and Electronic Engineering, University of Sheffield,
                     Sheffield, U.K.
			}
   \IEEEauthorblockA{\IEEEauthorrefmark{2}%
                     INRIA,
                     Centre Inria d'Universit\'e C\^ote d'Azur,
                     Sophia Antipolis, France.
                     }
   \IEEEauthorblockA{\IEEEauthorrefmark{3}%
                     ECE Dept., Princeton University, Princeton,
                     08544 NJ, USA.}
   \IEEEauthorblockA{\IEEEauthorrefmark{5}%
                     GAATI, Universit\'e de la Polyn\'esie Fran\c{c}aise,
                     Faaa, French Polynesia.}
}
%

\maketitle
\begin{abstract}
    The dual formulation of empirical risk minimization with $f$-divergence regularization (ERM-$f$DR) is introduced. The solution of the dual optimization problem to the ERM-$f$DR is connected to the notion of normalization function introduced as an implicit function. This dual approach leverages the Legendre-Fenchel transform and the implicit function theorem to provide a nonlinear ODE expression to the normalization function. Furthermore, the nonlinear ODE expression and its properties provide a computationally efficient method to calculate the normalization function of the ERM-$f$DR solution under a mild condition.
\end{abstract}

%
%
\section{Introduction}
\label{sec:introduction}
Empirical risk minimization (ERM)~\cite{vapnik1964perceptron, vapnik1974theory, vapnik1992principles, vapnik1993local, krzyzak1996nonparametric, zou2009generalization} is often posed as an optimization problem regularized by a \emph{statistical distance} between the probability measure to be optimized and a given reference measure~\cite{esposito2021generalization, wang2019information, dalalyan2024user, li2025regularization, peng2025information, futami2023informationtheoretic, huang2021stochastic}.
A well-studied case for such statistical distance is the relative entropy, which leads to the celebrated ERM with relative entropy regularization (ERM-RER) studied, for instance, in~\cite{raginsky2016information, russo2019much, perlazaISIT2022, daunas2024TITAsymmetry, perlaza2024ERMRER, aminian2021exact, bu2020tightening, zou2024WorstCase}.
Other works address the general case for $f$-divergences, known as ERM with $f$-divergence regularization (ERM-$f$DR)~\cite{teboulle1992entropic, beck2003mirror, alquier2021non, perlazaISIT2024a}.
The discrete case of ERM-$f$DR is explored in~\cite{teboulle1992entropic} and~\cite{beck2003mirror}, while more general settings are covered in~\cite{alquier2021non} and~\cite{perlazaISIT2024a}.
Recent results have shown that $f$-divergence regularization can improve the robustness of learning algorithms in the context of distributionally robust optimization (DRO)~\cite{wei2021optimizing, liu2023smoothed,namkoong2016stochastic,hu2013kulback}.

However, a main challenge in both the theory and practice of ERM-$f$DR is that the solution is known only up to a normalization factor~\cite{teboulle1992entropic, beck2003mirror, alquier2021non}.
For many $f$-divergences, a closed-form expression for the normalization factor is not known.
This hinders sampling methods like Markov Chain Monte Carlo (MCMC) \cite{hastings1970monte}, where it appears in the likelihood ratio, which is needed for valid transition probabilities. In the case of rejection sampling~\cite{gilks1992adaptive}, the absence of such a constant prevents defining an efficient proposal distribution.
Even when the normalization factor has a known expression, such as the log-partition function in the case of relative entropy regularization~\cite{perlazaISIT2023b}, it can be intractable for general loss functions. More broadly, in ERM-$f$DR problems, computing this normalization factor is especially difficult because it requires evaluating the empirical risk across all models in the support of the reference measure, a task that is $\#P$-hard~\cite{bulatov2005complexity, bulatov2013complexity, mcquillan2013computational}. 

This paper addresses this challenge by formalizing the concept of the normalization function introduced in~\cite{perlazaISIT2024a} and deriving the dual optimization problem for ERM-$f$DR. %
The benefits of studying the dual optimization problem and its properties are twofold. First, insights for computing the normalization factor are obtained, and second, a characterization of the normalization function for ERM-$f$DR problems is presented. 
This characterization is obtained by leveraging tools from convex analysis, notably the Legendre-Fenchel transform~\cite{rockafellar1970conjugate,boyd2004convex}, and the implicit function theorem~\cite{oswaldo2013TIFT}.
 
%
%
\section{Preliminaries}

Let $\Omega$ be an arbitrary subset of $\reals^{d}$, with $d \in \ints$, and let $\BorSigma{\Omega}$ denote the Borel $\sigma$-field on $\Omega$. The set of probability measures that can be defined upon the measurable space $\left(\Omega, \BorSigma{\Omega} \right)$ is denoted by~$\bigtriangleup(\Omega)$.
Given a probability measure $Q \in \bigtriangleup(\Omega)$ the set exclusively containing the probability measures in $\bigtriangleup(\Omega)$ that are absolutely continuous with respect to $Q$ is denoted by $\bigtriangleup_{Q}(\Omega)$. That is,
%
$\bigtriangleup_{Q}(\Omega) \triangleq \{P\in \bigtriangleup(\Omega): P \ll Q \}$,
%
where the notation $P \ll Q$ stands for the measure $P$ being absolutely continuous with respect to the measure $Q$.
The Radon-Nikodym derivative of the measure $P$ with respect to $Q$ is denoted by $\frac{\diff P}{\diff Q}:\Omega\rightarrow [0,\infty)$.

Using this notation, an $f$-divergence is defined as follows.
\begin{definition}[$f$-divergence~\cite{csiszar1967information}]
\label{Def_fDivergence}
Let $f:[0,\infty)\rightarrow \reals$ be a convex function with $f(1)=0$ and $f(0) \triangleq \lim_{x\rightarrow 0^+}f(x)$.
Let $P$ and $Q$ be two probability measures on the same measurable space, with $P \ll Q$.
The $f$-divergence of $P$ with respect to $Q$, denoted by $\KLf{P}{Q}$, is
\begin{equation}
\label{EqD_f}
\KLf{P}{Q} \triangleq \int f(\frac{\diff P}{\diff Q}(\omega)) \diff Q(\omega).
\end{equation}
\end{definition}

In the case in which the function $f$ in~\eqref{EqD_f} is continuous and differentiable, the derivative of the function~$f$ is denoted~by
$\dot{f}: (0, +\infty) \to \reals$.
If the inverse of the function~$\dot{f}$ exists, it is denoted by 
\begin{equation}
\label{EqDefInvDiffF}
\dot{f}^{-1}: \reals \to (0, +\infty). 
\end{equation}

%
%
\section{The Learning Problem}
\label{sec:ERMfDR}
%
Let~$\set{M}$,~$\set{X}$ and~$\set{Y}$, with~$\set{M} \subseteq \reals^{d}$ and~$d \in \ints$, be sets of \emph{models}, \emph{patterns}, and \emph{labels}, respectively.
A pair $(x,y) \in \mathcal{X} \times \mathcal{Y}$ is referred to as a \emph{labeled pattern} or \emph{data point}, and a \emph{dataset} is represented by the tuple $((x_1, y_1), (x_2, y_2), \ldots,(x_n, y_n))\in ( \set{X} \times \set{Y} )^n$.
Let the function~$h: \set{M} \times \mathcal{X} \rightarrow \mathcal{Y}$ be such that the label assigned to a pattern $x \in \set{X}$ according to the model $\thetav \in \set{M}$ is $h(\thetav,x)$.
Then, given a dataset
\begin{equation}
\label{EqTheDataSet}
\vect{z} = \big((x_1, y_1), (x_2, y_2 ), \ldots, (x_n, y_n )\big)  \in ( \set{X} \times \set{Y} )^n,
\end{equation}
the objective is to obtain a model $\thetav \in \set{M}$, such that, for all $i \inCountK{n}$, the label assigned to the pattern $x_i$, which is $h(\thetav,x_i)$, is ``close'' to the label $y_i$.
This notion of ``closeness'' is formalized by the function
    $\ell: \set{Y} \times \set{Y} \rightarrow [0, +\infty)$,
%
such that the loss or risk induced by choosing the model $\thetav \in \set{M}$  with respect to the labeled pattern $(x_i, y_i)$, with $i\inCountK{n}$, is $\ell(h(\thetav,x_i),y_i)$.
The risk function $\ell$ is assumed to be nonnegative and to satisfy $\ell( y, y ) = 0$, for all $y\in\set{Y}$.
The \emph{empirical risk} induced by a model $\vect{\theta}$ with respect to the dataset $\vect{z}$ in~\eqref{EqTheDataSet} is determined by the function \mbox{$\foo{L}_{\vect{z}}\!:\! \set{M} \rightarrow [0, +\infty)$}, which satisfies
\begin{IEEEeqnarray}{rcl}
\label{EqLxy}
\foo{L}_{\vect{z}} (\vect{\theta} )  &\ = \ &
\frac{1}{n}\smash{\sum_{i=1}^{n}} \ell ( h(\vect{\theta}, x_i), y_i ).
\end{IEEEeqnarray}%
%
The expectation of the empirical risk $\mathsf{L}_{\vect{z}} (\vect{\theta} )$ in~\eqref{EqLxy}, when~$\vect{\theta}$~is sampled from a probability measure $P \in \bigtriangleup(\set{M})$, is determined by the functional $\mathsf{R}_{\dset{z}}: \bigtriangleup(\set{M}) \rightarrow  [0, +\infty)$, such~that%
\begin{equation}
\label{EqRxy}
\foo{R}_{\dset{z}}( P ) = \int \foo{L}_{ \dset{z} } ( \thetav )  \diff P(\thetav).
\end{equation}
%

The ERM-$f$DR problem is parametrized by a probability measure $Q \in \bigtriangleup(\set{M})$, a positive real $\lambda$, and a function $f:[0,\infty)\to\reals$ that satisfies the conditions in Definition~\ref{Def_fDivergence}.
The measure $Q$ is referred to as the \emph{reference measure} and $\lambda$ as the \emph{regularization factor}.

Given the dataset~$\dset{z} \in (\set{X} \times \set{Y})^n$ in~\eqref{EqTheDataSet}, the \mbox{ERM-$f$DR} problem, with parameters~$Q$,~$\lambda$ and $f$, is given by the following optimization problem
\begin{IEEEeqnarray}{rcl}
\label{EqOp_f_ERMRERNormal}
\min_{P \in \bigtriangleup_{Q}(\set{M})} & \quad \foo{R}_{\dset{z}} ( P ) + \lambda \KLf{P}{Q},
\end{IEEEeqnarray}
where the functional $\foo{R}_{\dset{z}}$ is defined in~\eqref{EqRxy}.
The set of solutions to~\eqref{EqOp_f_ERMRERNormal} is the singleton $\{Q\}$ in the case in which for all $\thetav \in \supp Q$,  $\foo{L}_{\dset{z}}(\thetav) = c$, for some $c > 0$.
This distinction is mathematically significant but can be ignored in practice, as it arises only when $\foo{R}_{\dset{z}}(P)$ in~\eqref{EqRxy} is constant for all measures $P$. In order to avoid the above case, the notion of separable empirical risk functions~\cite[Definition 5]{perlaza2024ERMRER} is adopted.

The solution to the ERM-$f$DR problems in~\eqref{EqOp_f_ERMRERNormal} was first presented in \cite[Theorem~$1$]{perlazaISIT2024a} under the following assumptions: 
\begin{itemize}
\item[\namedlabel{assum:a}{$(a)$}] The function $f$ is strictly convex and twice differentiable;
\item[\namedlabel{assum:b}{$(b)$}] There exists a $\beta$ such that
\begin{subequations}
\label{EqfKrescConstrainAll}
\begin{equation}
\label{EqDefSetB}
\beta \in \left\lbrace t\in \reals: \forall \vect{\theta} \in \supp Q , 0 < \dot{f}^{-1} \left( -\frac{t + \foo{L}_{\vect{z}}(\thetav)}{\lambda} \right) \right\rbrace,
\end{equation}%
\vspace{-2mm}
and
\begin{IEEEeqnarray}{rCl}
\label{EqEqualToABigOne}
\int \dot{f}^{-1}(-\frac{\beta + \foo{L}_{\dset{z}}(\thetav)}{\lambda}) \diff Q(\thetav) & = & 1,
\IEEEeqnarraynumspace
\end{IEEEeqnarray}
where the function $\foo{L}_{\dset{z}}$ is defined in~\eqref{EqLxy}; and 
\item[\namedlabel{assum:c}{$(c)$}] The function $\foo{L}_{\dset{z}}$ in~\eqref{EqLxy} is separable with respect to the probability measure $Q$. 
\end{subequations}
\end{itemize}
%

\begin{theorem}[{\cite[Theorem~$1$]{perlazaISIT2024a}}]
\label{Theo_f_ERMRadNik}
Under Assumptions \ref{assum:a} and \ref{assum:b}, the solution to the optimization problem in~\eqref{EqOp_f_ERMRERNormal}, denoted by $\Pgibbs{P}{Q} \in \bigtriangleup_{Q}(\set{M})$, is unique, and for all $\thetav \in \supp Q$,%
\begin{equation}
\label{EqGenpdffDv}
\frac{\diff \Pgibbs{P}{Q}}{\diff Q} ( \thetav ) = \dot{f}^{-1}(-\frac{\beta + \foo{L}_{\dset{z}}(\thetav)}{\lambda}).
\end{equation}
\end{theorem}
%
The equality in~\eqref{EqGenpdffDv} can be written in terms of the \emph{normalization function}, introduced in~\cite{perlazaISIT2024a} and defined hereunder.
\begin{definition}[Normalization Function]
The normalization function of the problem in~\eqref{EqOp_f_ERMRERNormal}, denoted by
\begin{subequations}
\label{EqDefNormFunction}
\begin{equation}
\label{EqDefMapNormFunction}
N_{Q, \dset{z}}: \set{A}_{Q,\dset{z}} \rightarrow \set{B}_{Q,\dset{z}},
\end{equation}
with $\set{A}_{Q, \dset{z}}\subseteq (0,\infty)$ and $\set{B}_{Q,\dset{z}}\subseteq \reals$, is such that for all $\lambda \in \set{A}_{Q, \dset{z}}$,
\begin{IEEEeqnarray}{rCl}
\label{EqReasonNisNormFoo}
\int \dot{f}^{-1}\left(-\frac{N_{Q, \dset{z}}(\lambda) + \foo{L}_{\dset{z}}(\thetav)}{\lambda}\right) \diff Q(\thetav) = 1.
\IEEEeqnarraynumspace
\end{IEEEeqnarray}
\end{subequations}
\end{definition}
The set $\set{A}_{Q, \dset{z}}$ in~\eqref{EqDefMapNormFunction} contains all the regularization factors~$\lambda$ for which Assumption \ref{assum:b} is satisfied. More specifically, it contains the regularization factors $\lambda$ for which the problem in~\eqref{EqOp_f_ERMRERNormal} has a solution.
Furthermore, the equality in~\eqref{EqReasonNisNormFoo} justifies referring to the function $N_{Q, \dset{z}}$ as the \emph{normalization function}, as it ensures that the measure $\Pgibbs{P}{Q}$ in~\eqref{EqGenpdffDv} is a probability measure.
 
This section ends by highlighting that the probability measures $\Pgibbs{P}{Q}$ and $Q$ in~\eqref{EqGenpdffDv} are mutually absolutely continuous \cite[Corollary~$1$]{perlazaISIT2024a}.
This is important, as it ensures the Radon-Nikodym derivative of $Q$ with respect to $\Pgibbs{P}{Q}$ is well defined, a property that is used repeatedly throughout the proofs.

%
%
\section[ERM-fDR Dual Problem]{The ERM-$f$DR Dual Problem}
\label{sec:dual}
The duality principle \cite[Chapter~5]{boyd2004convex} enables the analysis of the optimization problem in~\eqref{EqOp_f_ERMRERNormal} by studying an alternative form, known as the dual problem.
In this section, this dual problem is derived using the Legendre-Fenchel transform~\cite{boyd2004convex}, which is defined below.
\begin{definition}[Legendre-Fenchel transform {\cite{boyd2004convex}}]
\label{DefLT_cnvxcnj}
Consider a function $f:\set{I}_1\rightarrow \set{I}_2$, with $\set{I}_i \subseteq \reals$, with $i \in \{1,2\}$. The Legendre-Fenchel transform of the function $f$, denoted by $f^{*}:\set{J} \rightarrow \reals$, is 
\begin{equation}
\label{EqDefLT_cnvxcnj}
	f^{*}(t) = \smash{\sup_{x\in \set{I}_1}}( tx- f(x)),
\end{equation}
with%
\begin{equation}
\label{EqDefJinLFT}
	\set{J} =  \{t \in \reals :f^{*}(t)<\infty\}.
\end{equation}
\end{definition}
Using this notation, consider the following problem 
\begin{IEEEeqnarray}{rcl}
\label{EqOp_f_ERMRERDual}
  \min_{\beta \in \reals} & \quad \lambda \int f^{*}\Big(-\frac{1}{\lambda}(\beta+\foo{L}_{\dset{z}}(\thetav))\Big)\diff Q(\thetav) + \beta,
\end{IEEEeqnarray}
where the real $\lambda$, the measure $Q$ and the function $f$ are those in~\eqref{EqOp_f_ERMRERNormal}; and the functions $\foo{L}_{\dset{z}}$ and $f^{*}$ are defined in~\eqref{EqLxy} and~\eqref{EqDefLT_cnvxcnj}, respectively.
The following theorem introduces the solution to the problem in~\eqref{EqOp_f_ERMRERDual}. 
\begin{theorem}
\label{Theo_dual_is_N}
Under Assumptions \ref{assum:a} and \ref{assum:b}, the solution to the optimization problem in~\eqref{EqOp_f_ERMRERDual} is $N_{Q,\dset{z}}(\lambda)$, where the function $N_{Q,\dset{z}}$ is defined in~\eqref{EqDefNormFunction}.
\end{theorem}
\begin{IEEEproof}
Let $G:\reals \to \reals$ be a function such that 
\begin{IEEEeqnarray}{rcl}
\label{EqDefDualCost}
  	G(\beta) &\ = \ & \lambda \int \Big(\!-\frac{1}{\lambda}(\beta+\foo{L}_{\dset{z}}(\thetav))\Big)\diff Q(\thetav) + \beta,
\end{IEEEeqnarray}
which is the objective function of the optimization problem in~\eqref{EqOp_f_ERMRERDual}. 
Note that from Assumption~\ref{assum:a} and Definition~\ref{DefLT_cnvxcnj}, the function $G$ in~\eqref{EqDefDualCost} is a strictly convex function, which satisfies
\begin{IEEEeqnarray}{rcl}
  	\!\!\!\frac{\diff}{\diff \beta}G(\beta) 
        &\ = \ &\! \frac{\diff}{\diff \beta}\Big(\lambda\! \int\! f^{*}\Big(\!-\!\frac{1}{\lambda}(\beta+\foo{L}_{\dset{z}}(\thetav))\!\Big)\diff Q(\thetav)\!+\! \beta\Big)\ \ \\
  	&\ = \ & -\! \int\! \dot{f^{*}}\Big(-\frac{1}{\lambda}(\beta+\foo{L}_{\dset{z}}(\thetav))\Big)\diff Q(\thetav) + 1,\label{EqDiffG_beta_s3}
\end{IEEEeqnarray}
%
where $\dot{f^{*}}$ is the derivative of the function $f^{*}$ in \eqref{EqOp_f_ERMRERDual}.
Let the solution to the optimization problem in~\eqref{EqDefDualCost} be denoted by $\widehat{\beta} \in \reals$ and note that the derivative of the function $G$ evaluated at $\widehat{\beta}$ is equal to zero, that is
\begin{IEEEeqnarray}{rcl}
  	\int \dot{f^{*}}\Big(-\frac{1}{\lambda}\big(\widehat{\beta}+\foo{L}_{\dset{z}}(\thetav)\big)\Big)\diff Q(\thetav) &\ = \ & 1.\label{EqDiffG_betaStar}
\end{IEEEeqnarray}
From~\cite[Corollary~$23.5.1$]{rockafellar1970conjugate} and Assumption~\ref{assum:a}, the following equality holds for all $t \in \set{J}$, with $\set{J}$ in~\eqref{EqDefJinLFT},
\begin{IEEEeqnarray}{rCCCl}
\label{EqDefLFTDiffF}
\frac{\diff}{\diff t}f^{*}(t) & = & \dot{f^{*}}(t)& = & \dot{f}^{-1}(t),
\end{IEEEeqnarray}
where the functions $\dot{f}^{-1}$ and $\dot{f^{*}}$ are defined in~\eqref{EqDefInvDiffF} and \eqref{EqDiffG_beta_s3}, respectively.
From~\eqref{EqDiffG_betaStar} and~\eqref{EqDefLFTDiffF}, it follows that
\begin{IEEEeqnarray}{rcl}
\label{EqConditionDualIntIs1}
  	\int \dot{f}^{-1}\Big(-\frac{1}{\lambda}\big(\widehat{\beta}+\foo{L}_{\dset{z}}(\thetav)\big)\Big)\diff Q(\thetav) &\ = \ & 1,
\end{IEEEeqnarray}
which combined with~\eqref{EqReasonNisNormFoo} and Assumption~\ref{assum:b} yields
\begin{IEEEeqnarray}{rcl}
  	 N_{Q,\dset{z}}(\lambda) &\ = \ & \widehat{\beta}.
\end{IEEEeqnarray}
Note that $\hat{\beta}$ is unique as it is the minimum of a strictly convex function.
This observation completes the proof.
\end{IEEEproof}

The following lemma establishes that the problem in~\eqref{EqOp_f_ERMRERDual} is the dual problem to the ERM-$f$DR problem in~\eqref{EqOp_f_ERMRERNormal} and characterizes the difference between their optimal values, which is often referred to as duality gap \cite[Section 8.3]{luenberger1964observing}.
\begin{lemma}
\label{lemm_ZeroDualGap}
Under Assumptions~\ref{assum:a} and~\ref{assum:b}, the optimization problem in~\eqref{EqOp_f_ERMRERDual} is the dual problem to the ERM-$f$DR problem in~\eqref{EqOp_f_ERMRERNormal}.
Moreover, the duality gap is zero. 	
\end{lemma}
\begin{IEEEproof}
	Under Assumption~\ref{assum:a} and \cite[Section~$3.3.2$]{boyd2004convex}, it can be verified that for all $t \in \set{J}$, with $\set{J}$ in~\eqref{EqDefJinLFT}, the function $f^{*}$ in~\eqref{EqDefLT_cnvxcnj} satisfies 
	\begin{IEEEeqnarray}{rcl}
	\label{EqLFT_cool_equality}
  	f^{*}(t) &\ = \ & t\dot{f^{*}}(t)-f\big(\dot{f^{*}}(t)\big),
  	\IEEEeqnarraynumspace
	\end{IEEEeqnarray}
	where the function $\dot{f}^{*}$ is the same as in~\eqref{EqDefLFTDiffF}.
	From Assumption~\ref{assum:a} and~\eqref{EqDefLFTDiffF}, the Radon-Nikodym derivative $\frac{\diff \Pgibbs{P}{Q}}{\diff Q}$ in~\eqref{EqGenpdffDv} satisfies for all $\thetav \in \supp Q$,
	\begin{IEEEeqnarray}{rCl}
	\label{EqGenpdffDvLFT}
  	\frac{\diff \Pgibbs{P}{Q}}{\diff Q}(\thetav) &\ = \ & \dot{f^{*}}(-\frac{N_{Q, \dset{z}}(\lambda) + \foo{L}_{\dset{z}}(\thetav)}{\lambda}),
  	\IEEEeqnarraynumspace
	\end{IEEEeqnarray}
	where the functions $\foo{L}_{\dset{z}}$ and $N_{Q, \dset{z}}$ are defined in~\eqref{EqLxy} and \eqref{EqDefNormFunction}, respectively.
	Then, from~\eqref{EqLFT_cool_equality} and~\eqref{EqGenpdffDvLFT}, with some algebraic manipulations, it holds that for all $\thetav \in \supp Q$,
	\begin{IEEEeqnarray}{rcl}
	\IEEEeqnarraymulticol{3}{l}{
  	\!\!\foo{L}_{\dset{z}}(\thetav)+\lambda f\bigg(\frac{\diff \Pgibbs{P}{Q}}{\diff Q}(\thetav)\!\!\bigg)\frac{\diff Q}{\diff \Pgibbs{P}{Q}}(\thetav)  
  	}\nonumber \\
  	\!\!& = & -\lambda f^{*}\!\Big(\!-\frac{ N_{Q, \dset{z}}(\lambda)\! +\! \foo{L}_{\dset{z}}(\thetav)}{\lambda}\Big)\frac{\diff Q}{\diff \Pgibbs{P}{Q}}(\thetav)\!-\! N_{Q, \dset{z}}(\lambda).\quad \ \ \label{EqPfDualLTArrange}
	\end{IEEEeqnarray}
	Taking the expectation in both sides of~\eqref{EqPfDualLTArrange} with respect to the probability measure $\Pgibbs{P}{Q}$ in~\eqref{EqGenpdffDv} yields
	\begin{IEEEeqnarray}{rcl}
	\IEEEeqnarraymulticol{3}{l}{
  	\foo{R}_{\dset{z}}(\Pgibbs{P}{Q})+\lambda \Divf{\Pgibbs{P}{Q}}{Q}  
  	}\nonumber \\
  	&\ = \ &\!- \lambda\! \int\! f^{*}\Big(\!-\frac{1}{\lambda}(N_{Q, \dset{z}}(\lambda)\!+\!\foo{L}_{\dset{z}}(\thetav))\Big)\diff Q(\thetav)\!-\! N_{Q, \dset{z}}(\lambda).\qquad \label{EqPfDualLTExpect}
	\end{IEEEeqnarray}
	Using Theorem~\ref{Theo_f_ERMRadNik} and Theorem~\ref{Theo_dual_is_N} in the left-hand and right-hand sides of~\eqref{EqPfDualLTExpect}, respectively, yields
	\begin{IEEEeqnarray}{rcl}
	\IEEEeqnarraymulticol{3}{l}{
  	\min_{P\in\bigtriangleup_{Q}(\set{M})}\foo{R}_{\dset{z}}(P)+\lambda \Divf{P}{Q}  
  	}\nonumber \\
  	&\ = \ & \max_{\beta\in \reals}- \lambda\! \int f^{*}\Big(-\frac{1}{\lambda}(\beta+\foo{L}_{\dset{z}}(\thetav))\Big)\diff Q(\thetav)-\beta.  \ \  \label{EqPfDualLTOpt}
	\end{IEEEeqnarray}
	The claim that the optimization problem in~\eqref{EqOp_f_ERMRERDual} is the dual to the ERM-$f$DR problem in~\eqref{EqOp_f_ERMRERNormal} follows from~\eqref{EqPfDualLTOpt} and \cite[Theorem~$1$, Section~$8.4$]{luenberger1997bookOptimization}. The zero duality gap is established by the equality in~\eqref{EqPfDualLTOpt}, which completes the proof.
	\end{IEEEproof}
	
	The zero duality gap in Lemma~\ref{lemm_ZeroDualGap} ensures strong duality, which implies that the optimal value of the dual variables recovers the primal optimal measure.
	However, solving the dual problem in~\eqref{EqOp_f_ERMRERDual} remains equally challenging, as the solution must satisfy Assumption~\ref{assum:b} in Theorem~\ref{Theo_f_ERMRadNik}. Thus, unless the normalization function $N_{Q, \dset{z}}$ in \eqref{EqDefNormFunction} is explicitly characterized, computing $N_{Q, \dset{z}}(\lambda)$ involves a search over the real line to obtain the value that satisfies \eqref{EqConditionDualIntIs1}.

%
%
\section{Analysis of the Normalization Function}
\label{sec:analysisRegFact}
\subsection{Characterization and Properties}
The purpose of this section is to characterize the function $N_{Q, \dset{z}}$ and the sets $\set{A}_{Q, \dset{z}}$ and $\set{B}_{Q, \dset{z}}$ in~\eqref{EqDefNormFunction}.
Given a real~$\delta\in [0, \infty)$, consider the Rashomon set $\set{L}_{\dset{z}}(\delta)$, which is defined as follows 
	$\smash{\set{L}_{\dset{z}}(\delta) \triangleq \{\thetav \in \set{M}: \foo{L}_{\dset{z}}(\thetav) \leq \delta \}}$.
%
%
Consider also the real numbers $\delta^\star_{Q, \dset{z}}$ and $\lambda^\star_{Q, \dset{z}}$ defined as follows
%
\begin{IEEEeqnarray}{rCl}
\label{EqDefDeltaStar}
\delta^\star_{Q, \dset{z}} &\triangleq & \inf \{\delta \in [0, \infty): Q(\set{L}_{\dset{z}}(\delta))>0\},
\end{IEEEeqnarray}%
and%
\vspace{-3mm}
\begin{IEEEeqnarray}{rCl}
	 \label{EqDefLambdaStar}
		\smash{\lambda^{\star}_{Q, \dset{z}}} &\triangleq & \smash{\inf\set{A}_{Q,\dset{z}}}.
\end{IEEEeqnarray}
Using this notation, the following theorem introduces one of the main properties of the function $N_{Q, \dset{z}}$.

\begin{theorem}
\label{theo_InfDevKfDR}
The function~$N_{Q, \dset{z}}$ in~\eqref{EqDefNormFunction} is strictly increasing and continuous within the interior of $\set{A}_{Q, \dset{z}}$ in~\eqref{EqDefMapNormFunction}. Furthermore,  for all $\lambda \in \set{A}_{Q, \dset{z}}$,
\begin{IEEEeqnarray}{rcl}
\label{EqNQzExplisit}
	N_{Q, \dset{z}}(\lambda) 
	&\ = \ & \smash{\lambda \frac{\diff }{\diff \lambda}N_{Q, \dset{z}}(\lambda) - \foo{R}_{\dset{z}}(P_N)},
\end{IEEEeqnarray}
where the probability measure $P_N \in \bigtriangleup_{Q}(\set{M})$ satisfies for all $\thetav \in \supp Q$,
\begin{equation}
	\frac{\diff P_N}{\diff Q}\!(\thetav)\!  =\! 
	\smash{\frac{1}{\ddot{f}(\!\frac{\diff \Pgibbs[\dset{z}][a]{P}{Q}}{\diff Q}(\thetav)\!)}  \bigg(\int \frac{1}{\ddot{f}(\!\frac{\diff \Pgibbs[\dset{z}][a]{P}{Q}}{\diff Q}(\nuv)\!\!)}\! \diff Q (\nuv)\!\!\bigg)^{-1}\!\!\!\!\!\!\!.\quad}
	\label{EqPofLambda}
\end{equation}
\end{theorem}
\subsection{Proof of Theorem~\ref{theo_InfDevKfDR}}
\label{AppProofLemmaInfDevKDivf}
The proof is divided into two parts.
The first part leverages the properties of the function $f$ under Assumption~\ref{assum:a} in Theorem~\ref{Theo_f_ERMRadNik} to prove that the normalization function \mbox{$N_{Q, \dset{z}}:\set{A}_{Q, \dset{z}} \rightarrow \set{B}_{Q, \dset{z}}$} in~\eqref{EqDefNormFunction} is strictly increasing.
The second part proves the continuity of the function $N_{Q, \dset{z}}$ in~\eqref{EqDefNormFunction}.

The first part is as follows.
Given a pair $(a,b) \in \set{A}_{Q, \dset{z}}\times\set{B}_{Q, \dset{z}}$, with $\set{A}_{Q, \dset{z}}$ and $\set{B}_{Q,\dset{z}}$ in~\eqref{EqDefNormFunction}, assume that
\begin{equation}
\label{EqProofKrescalingL1}
N_{Q, \dset{z}}(a) = b.
\end{equation}
This implies that
\begin{equation}
1 
 \!= \!\! \int\! \frac{\diff \Pgibbs[\dset{z}][b]{P}{Q}}{\diff Q}(\thetav) \diff Q(\thetav)
\label{Eq_ProofLambdaIsTheInvOfKbar_s1}
 = \!\! \int\! \dot{f}^{-1}\Big(\!-\frac{1}{a}(b+\foo{L}_{\dset{z}}(\thetav))\!\Big)\!\diff Q(\thetav).
\end{equation}
Note that the inverse $\dot{f}^{-1}$ exists from the fact that $f$ is strictly convex, which implies that $\dot{f}$ is a strictly increasing function. Hence, $\dot{f}^{-1}$ is also a strictly increasing function in $\set{B}_{Q, \dset{z}}$~\cite[Theorem~$5.6.9$]{bartle2000introduction}.
Moreover, from the assumption that $f$ is strictly convex and differentiable, it holds that $\dot{f}$ is continuous~\cite[Proposition 5.44]{douchet2010Analyse}. This implies that $\dot{f}^{-1}$ is continuous.
From~\cite[Lemma~A.3]{InriaRR9594}, the function $\dot{f}^{-1}$ is strictly increasing such that for all $b \in \set{B}_{Q, \dset{z}}$ and for all $\thetav \in \supp Q$, it holds that
\begin{IEEEeqnarray}{rCl}
\label{Eq_ProofFinitenes_pf}	
\dot{f}^{-1}\Big(\!-\frac{1}{a}(b\!+\!\foo{L}_{\dset{z}}(\thetav))\Big)
& \leq & \dot{f}^{-1}\Big(\!-\frac{1}{a}(b\!+\!\delta^\star_{Q, \dset{z}})\Big)\!<\!\infty,\quad
\end{IEEEeqnarray}%
with $\delta^\star_{Q, \dset{z}}$ defined in~\eqref{EqDefDeltaStar} and equality holds if and only if $ \foo{L}_{\dset{z}}(\thetav)  = \delta^\star_{Q, \dset{z}}$.
Then, from~\eqref{Eq_ProofFinitenes_pf} and $\set{A}_{Q, \dset{z}} \subseteq (0,\infty)$, which implies $a > 0$, it follows that
\begin{IEEEeqnarray}{rCl}
\int\! \dot{f}^{-1}\Big(\!\!-\frac{1}{a}(b\!+\!\foo{L}_{\dset{z}}(\thetav))\Big)\diff Q(\thetav)
& < & \!\infty. \quad\label{Eq_ProofKbarNoLambdaIsFinite_pf_s3}
\end{IEEEeqnarray}
Note that from \cite[Lemma~A.5]{InriaRR9594} and the assumption in \eqref{Eq_ProofLambdaIsTheInvOfKbar_s1} for all $(a_1, a_2) \in \set{A}_{Q, \dset{z}}^2$, such that $a_1 < a < a_2$, it holds that
\begin{equation}
	\!\int\!\! \dot{f}^{-1}\Big(\!-\frac{\foo{L}_{\dset{z}}(\thetav)\!+\! b}{a_1} \Big)\! \diff Q(\thetav)\! < \! 1\!  < \!\! \int\!\! \dot{f}^{-1}\Big(\!-\frac{\foo{L}_{\dset{z}}(\thetav)\!+\!b}{a_2}\Big)\! \diff Q(\thetav).
\end{equation}
Then, for $N_{Q, \dset{z}}(a_i)$, with $i\in \{1,2\}$, to satisfy,
\begin{IEEEeqnarray}{rCl}
	\int  \dot{f}^{-1}(-\frac{1}{a_i}(\foo{L}_{\dset{z}}(\thetav) + N_{Q, \dset{z}}(a_i))) \diff Q(\thetav) & = & 1,
\end{IEEEeqnarray}%
under the assumption that $a_1 < a < a_2$, it must hold that $N_{Q, \dset{z}}(a_1)$ and $N_{Q, \dset{z}}(a_2)$ satisfy%
\begin{IEEEeqnarray}{rCCCl}
\label{Eq_ProofKbarStricIncreaseGam1vs2}
N_{Q, \dset{z}}(a_1)  & < & b & < & 	N_{Q, \dset{z}}(a_2).
\end{IEEEeqnarray}
This implies that the function $N_{Q, \dset{z}}$ in~\eqref{EqDefNormFunction} is strictly increasing.
Similarly, from \cite[Lemma~A.5]{InriaRR9594} and the assumption in \eqref{Eq_ProofLambdaIsTheInvOfKbar_s1} for all $(b_1, b_2) \in \set{B}_{Q, \dset{z}}^2$, such that $b_1 < b < b_2$, it holds that
\begin{equation}
	\!\int\!\! \dot{f}^{-1}\Big(\!\!-\frac{\foo{L}_{\dset{z}}(\thetav)\! +\! b_1}{a} \Big)\! \diff Q(\thetav)\! > \! 1\!  > \!\! \int\!\! \dot{f}^{-1}\Big(\!\!-\frac{\foo{L}_{\dset{z}}(\thetav)\! +\! b_2}{a}\Big)\! \diff Q(\thetav).
\end{equation}
Then, for the $a_i$, with $i \in \{1,2\}$ to satisfy,
\begin{IEEEeqnarray}{rCl}
	\int  \dot{f}^{-1}(-\frac{1}{a_i}(\foo{L}_{\dset{z}}(\thetav) + b_i)) \diff Q(\thetav) & = & 1,
\end{IEEEeqnarray}
under the assumption that $b_1 < b < b_2$, it holds that $a_1$ and $a_2$ satisfy%
\vspace{-2mm}
\begin{equation}
\label{Eq_ProofKbarStricIncreaseGam1vs2_2}
\smash{a_1   <  a  <  	a_2,}
\end{equation}
which implies that the function $N_{Q, \dset{z}}$ in~\eqref{EqDefNormFunction} is strictly increasing.
Furthermore, from~\eqref{Eq_ProofKbarStricIncreaseGam1vs2} and~\eqref{Eq_ProofKbarStricIncreaseGam1vs2_2} the function $N_{Q, \dset{z}}$ maps one to one the elements of $\set{A}_{Q, \dset{z}}$ into $\set{B}_{Q, \dset{z}}$, which implies it is bijective. Thus, the inverse $N^{-1}_{Q, \dset{z}}:\set{B}_{Q, \dset{z}}\to \set{A}_{Q, \dset{z}}$ is well-defined. 
This completes the proof of the first part. 

In the second part, the objective is to prove the continuity of the function $N_{Q, \vect{z}}$. To do so, an auxiliary function is introduced and proven to be continuous.
Under the assumptions \ref{assum:a}, \ref{assum:b} and \ref{assum:c} from Theorem~\ref{Theo_f_ERMRadNik}, the sets $\set{A}_{Q, \dset{z}}$ and $\set{B}_{Q, \dset{z}}$ in~\eqref{EqDefNormFunction} are non-empty and the real values: $\bar{a} = \sup \set{A}_{Q, \dset{z}}$; $\underline{a} =  \inf \set{A}_{Q, \dset{z}}$; $\bar{b} = \sup \set{B}_{Q, \dset{z}}$; and $\underline{b} = \inf \set{B}_{Q, \dset{z}}$, are such~that 
\begin{IEEEeqnarray}{rCCCl}
\label{EqDefABforF}
	\set{A} & = & (\underline{a},\bar{a}) \subseteq (0,\infty), \text{ and }
	\set{B} & = & (\underline{b},\bar{b}) \subseteq \reals.
\end{IEEEeqnarray}
Let the function $F: \set{A}\times \set{B} \to (0,\infty)$ be
\begin{IEEEeqnarray}{rcl}
\label{EqkWAWALem5}
F(a,b) & = &  \int \dot{f}^{-1}(-\frac{b + \foo{L}_{\dset{z}}(\thetav)}{a}) \diff Q (\vect{\theta})-1.
\end{IEEEeqnarray}
The first step is to prove that the function $F$ in~\eqref{EqkWAWALem5} is continuous on the sets $\set{A}$ and $\set{B}$ defined in~\eqref{EqDefABforF}, respectively. This is proved by showing that $F$ always exhibits a limit in $\set{A}$ and $\set{B}$, and that limit is the same as the function evaluated at that value.
Then, for all $(a,b) \in \set{A}\times \set{B}$ and for all $\vect{\theta} \in \supp Q$, the function $\dot{f}^{-1}$ satisfies the bound in \eqref{Eq_ProofFinitenes_pf}.
Hence, from~\cite[Corollary~$24.5.1$]{rockafellar1970conjugate} the function $\dot{f}^{-1}$ is continuous, such that for all $(a, b) \in \set{A}\times\set{B}$, it holds that
\begin{IEEEeqnarray}{rcl}
\label{EqLimRNparB}
\lim_{a \to \lambda}\lim_{b \to \beta} \dot{f}^{-1}(\frac{-b - \foo{L}_{\dset{z}}(\thetav)}{a}) & = &  \dot{f}^{-1}(\frac{-\beta - \foo{L}_{\dset{z}}(\thetav)}{\lambda}).\quad
\end{IEEEeqnarray}
Note that the function $F$ in~\eqref{EqkWAWALem5} is bounded as a consequence of \eqref{Eq_ProofFinitenes_pf}, and thus, from the \emph{dominated convergence theorem}~\cite[Theorem~1.6.9]{ash2000probability}, the following limits exist and~satisfy
\begin{IEEEeqnarray}{rcl}
\label{EqLimContFbeta}
\smash{\lim_{b \to \beta} F(a,b) }
& = &\smash{ F(\beta,a)},
\end{IEEEeqnarray}%
and\vspace{-2mm}%
\begin{IEEEeqnarray}{rcl}
\smash{\lim_{a \to \lambda} F(a,b) }
& = & \smash{F(\lambda,b)},\label{EqLimContFlambda}
\end{IEEEeqnarray}
which proves that the function $F$ in~\eqref{EqkWAWALem5} is continuous in $\set{A}$ and $\set{B}$.
The proof continues by noting that from the definition of $\set{A}$ and $\set{B}$ in~\eqref{EqDefABforF} there exists at least one point $(\lambda, \beta) \in \set{A}\times\set{B}$, such that $(\lambda, \beta) \in \set{A}_{Q,\dset{z}}\times \set{B}_{Q,\dset{z}}$, which implies that
\begin{IEEEeqnarray}{rcl}
\label{EqRootOfF}
 F(\lambda,\beta) & = & 0.
\end{IEEEeqnarray}
Note that from~\eqref{EqLimContFbeta} and~\eqref{EqLimContFlambda} the function $F$ is continuous and thus the partial derivative of $F$ satisfy
\begin{IEEEeqnarray}{rcl}
 \label{EqDeffaFab}
 \frac{\partial}{\partial a}F(a,b) 
 	& = &\! \int\! \frac{b + \foo{L}_{\dset{z}}(\thetav)}{a^{2}}\frac{1}{\ddot{f}(\dot{f}^{-1}(-\frac{b + \foo{L}_{\dset{z}}(\thetav)}{a}\!)\!)} \diff Q (\vect{\theta}),\qquad \label{EqDeffaFab_s4}
\end{IEEEeqnarray}
where~\eqref{EqDeffaFab_s4} follows from \cite[Lemma~A.2]{InriaRR9594}; and
\begin{IEEEeqnarray}{rcl}
 \label{EqDeffbFab}
 \frac{\partial}{\partial b}F(a,b) 
 	& = & \int -\frac{1}{a}\frac{1}{\ddot{f}(\dot{f}^{-1}(-\frac{b + \foo{L}_{\dset{z}}(\thetav)}{a}))} \diff Q (\vect{\theta}),\label{EqDeffbFab_s4}
\end{IEEEeqnarray}
where~\eqref{EqDeffbFab_s4} follows from \cite[Lemma~A.2]{InriaRR9594}.
Then, from the \emph{implicit function theorem} presented in~\cite[Theorem~$4$]{oswaldo2013TIFT}, the function~$N_{Q, \dset{z}}$ exists and is unique in the open interval $\set{A}$ with $\set{A}$ in~\eqref{EqDefABforF} and for all $a \in \set{A}$ satisfies that
\begin{equation}
\smash{N_{Q, \dset{z}}(a)  =  b,} 
\end{equation}
such that
%
$F(a,N_{Q, \dset{z}}(a)) =   0$,
which completes the proof of continuity for the normalization function $N_{Q, \dset{z}}$.
Additionally, from~\cite[Theorem~$4$]{oswaldo2013TIFT} it follows that 
\begin{IEEEeqnarray}{rcl}
\!\!\frac{\diff }{\diff a}N_{Q, \dset{z}}(a)
& = & - \! (\!\frac{\partial}{\partial b}F(a,N_{Q, \dset{z}}(a))\!)^{-1}\!\!\frac{\partial}{\partial a}\!F(a,N_{Q, \dset{z}}(a)\!), \qquad\label{EqProofNzFuncDef_s1}\\
& = & \frac{N_{Q, \dset{z}}(a)}{a}+ \displaystyle\int \frac{\foo{L}_{\dset{z}}(\thetav)}{a}g_a(\thetav)\diff Q (\thetav),\label{EqProofNzFuncDef_s7}
\end{IEEEeqnarray}
with the function $g_a:\set{M} \to \reals$, such that for all $\thetav \in \supp Q$,
\begin{IEEEeqnarray}{rcl}
\label{EqRNnewRm}
	g_a(\thetav) & = & 
	\frac{\diff P_N}{\diff Q}(\thetav),
\end{IEEEeqnarray}
with $\frac{\diff P_N}{\diff Q}$ defined in \eqref{EqPofLambda}.
Note that from the assumption that $f$ is strictly convex and twice differentiable, the derivative $\dot{f}$ is increasing, and the second derivative $\ddot{f}$ is positive for all $\thetav \in \supp Q$. Also, the denominator of the fraction is the integral of the reciprocal of $\ddot{f}\left( \frac{\diff \Pgibbs[\dset{z}][a]{P}{Q}}{\diff Q}(\nuv) \right)$ with respect to the measure $Q$. This term serves as a normalization constant ensuring that the resulting function is a proper probability density such that
%
	$\int g_a(\thetav) \diff Q (\thetav) =  1$.
%
Therefore, the function $g_a$ in~\eqref{EqRNnewRm} can be interpreted as the Radon-Nikodym derivative of a new probability measure $P^{(a)}$, parametrized by the regularization factor $a$ with respect to $Q$. Specifically, if we define a measure $P^{(a)}$ such that for any set $\set{A} \in \field{F}_{\set{M}}$,
\begin{IEEEeqnarray}{rcl}
\label{EqGaIsMeasureP}
	P^{(a)}(\set{A}) = \int_{\set{A}} g_a(\thetav) \diff Q(\thetav).
\end{IEEEeqnarray}
%
%
From~\eqref{EqProofNzFuncDef_s7} and~\eqref{EqGaIsMeasureP}, it follows that
\begin{IEEEeqnarray}{rcl}
	\label{EqODEformNQz}
	N_{Q, \dset{z}}(a) 
	& = & a \frac{\diff }{\diff a}N_{Q, \dset{z}}(a) - \foo{R}_{\dset{z}}\Big(P^{(a)}\Big),
\end{IEEEeqnarray}
with $\foo{R}_{\dset{z}}$ defined in~\eqref{EqRxy}.
This completes the proof of the derivative of the normalization function. \hfill\IEEEQED

 
The continuity and monotonicity exhibited by the function $N_{Q,\dset{z}}$ allow the following characterization of the set $\set{A}_{Q, \dset{z}}$.
%
\begin{lemma}
\label{lemm_fDR_kset}
The set $\set{A}_{Q, \dset{z}}$ in~\eqref{EqDefMapNormFunction} is either empty or an interval that satisfies 
%
 $			(\lambda^{\star}_{Q,\dset{z}}, \infty) \subseteq \set{A}_{Q, \dset{z}} \subseteq [\lambda^{\star}_{Q,\dset{z}}, \infty),
 $
with $\lambda^{\star}_{Q,\dset{z}}$ in \eqref{EqDefLambdaStar} and satisfies $\lambda^{\star}_{Q,\dset{z}} \geq 0$.
\end{lemma}
\begin{IEEEproof}
	The proof is presented in \cite[Appendix~B.1]{InriaRR9594}.
\end{IEEEproof}

Lemma~\ref{lemm_fDR_kset} highlights two facts. First, the set $\set{A}_{Q, \dset{z}}$ is a convex subset of positive reals. 
Second, if there exists a solution to the optimization problem in~\eqref{EqOp_f_ERMRERNormal} for some $\lambda>0$, then there exists a solution to such a problem when $\lambda$ is replaced by~$\bar{\lambda} \in(\lambda,\infty)$.

\subsection{Discussion of Results}
A major bottleneck in ERM-$f$DR solutions stems from the computational challenges associated with: $(i)$ evaluating expectations with respect to the prior $Q$, and $(ii)$ determining the value of $N_{Q, \dset{z}}(\lambda)$ in~\eqref{EqGenpdffDv} for implementing the resulting algorithms.
This subsection discusses the significance of the continuity and monotonicity properties of the normalization function 
$N_{Q, \dset{z}}$ in \eqref{EqDefNormFunction} and their practical implications for addressing challenge $(ii)$, under the assumption that expectations with respect to the prior $Q$ are computationally tractable.
To begin, note that the function $N_{Q, \dset{z}}$ induces a bijection between the sets $\set{A}_{Q, \dset{z}}$  and $\set{B}_{Q, \dset{z}}$, as a result of the monotonicity and continuity established in Theorem~\ref{theo_InfDevKfDR}.
For the cases in which no explicit expression for the normalization function is available, this bijection allows the ERM-$f$DR solution to be computed directly by selecting a normalization factor and using its inverse to determine the corresponding regularization parameter $\lambda$.
For example, this approach is used in \cite[Eq (26)]{daunas2024TITAsymmetry} for the reverse relative entropy.
More importantly, in cases in which neither the normalization function $N_{Q, \dset{z}}$ nor its inverse can be explicitly defined, as is the case for several $f$-divergences (see \cite{perlazaISIT2024a}); the monotonicity and continuity of $N_{Q, \dset{z}}$ allow replacing an exhaustive search over all real values with a root-finding algorithm to approximate the solution, ensuring the existence of the root in \eqref{EqRootOfF} and convergence to it.
Thus, such an algorithm reduces the number of evaluations of the expectation with respect to $Q$ to find such an approximation. The following Algorithm makes use of the above properties of the normalization function~$N_{Q,\dset{z}}$.

\begin{algorithm}[H]
\caption{Algorithm for $N_{Q,\dset{z}}(\lambda)$ via Root-finding}
\begin{algorithmic}[1]
\renewcommand{\algorithmicrequire}{\textbf{Input:}}
\renewcommand{\algorithmicensure}{\textbf{Output:}}
\REQUIRE $\foo{L}_{\dset{z}}: \set{M} \to [0,\infty)$, $Q$, $f: (0,\infty) \to \mathbb{R}$, $\lambda > 0$, $\delta^{\star}_{Q,\dset{z}}$, tolerance $\epsilon > 0$, max iterations $N_{\text{max}}\vphantom{1}$\vspace{1.5mm}
\ENSURE $\beta$ satisfying $\left|\int \dot{f}^{-1}\left(-\frac{\foo{L}_{\dset{z}}(\thetav)+\beta}{\lambda}\right) \diff Q(\thetav) - 1\right| \leq \epsilon$
\\ \textit{Initialisation}:
\STATE $b_{\text{low}} \gets \delta^{\star}_{Q,\dset{z}}-\lambda\dot{f}(0)$, $b_{\text{high}} \gets \lambda$, $b \gets \frac{1}{2}(b_{\text{low}} + b_{\text{high}})$
\STATE $n \gets 0$, $I \gets \int  \dot{f}^{-1}\left(-\frac{\foo{L}_{\dset{z}}(\thetav)+b}{\lambda}\right) \diff Q(\thetav)$
\\ \textit{Root-Finding Process}:
\WHILE{$|I - 1| > \epsilon$ \AND $n < N_{\text{max}}$}
    \IF{$I > 1$}
        \STATE $b_{\text{high}} \gets b$
    \ELSE
        \STATE $b_{\text{low}} \gets b$
    \ENDIF
    \STATE $b \gets \frac{1}{2}(b_{\text{low}} + b_{\text{high}})$
    \STATE $I \gets \int \dot{f}^{-1}\left(-\frac{\foo{L}_{\dset{z}}(\thetav)+b}{\lambda}\right) \diff Q(\thetav)$, $n \gets n + 1$
\ENDWHILE;  \textbf{return} $b$
\end{algorithmic}
\end{algorithm}
\section{Conclusions}\label{SecConclusions}

This paper establishes a connection between the ERM-$f$DR solution and the Legendre-Fenchel transform via the dual formulation in~\eqref{EqOp_f_ERMRERDual}. By analyzing the relationship between the primal and dual problems in~\eqref{EqOp_f_ERMRERNormal} and~\eqref{EqOp_f_ERMRERDual}, it is shown that the duality gap is zero. The identity arising from strong duality is leveraged to prove the continuity and monotonicity of the normalization function, enabling application of the implicit function theorem. This, in turn, yields a nonlinear ODE characterizing the normalization function. These properties are then used to design an algorithm that approximates the normalization function for a tractable reference measure and fixed regularization factor. This improves over searching across the real line when an explicit characterization is unavailable.

%
\IEEEtriggeratref{22}
\bibliographystyle{IEEEtranlink}
\bibliography{iEEEtranBibStyle.bib}

\begin{thebibliography}{10}
\providecommand{\url}[1]{#1}
\csname url@samestyle\endcsname
\providecommand{\newblock}{\relax}
\providecommand{\bibinfo}[2]{#2}
\providecommand{\BIBentrySTDinterwordspacing}{\spaceskip=0pt\relax}
\providecommand{\BIBentryALTinterwordstretchfactor}{4}
\providecommand{\BIBentryALTinterwordspacing}{\spaceskip=\fontdimen2\font plus
\BIBentryALTinterwordstretchfactor\fontdimen3\font minus \fontdimen4\font\relax}
\providecommand{\BIBforeignlanguage}[2]{{%
\expandafter\ifx\csname l@#1\endcsname\relax
\typeout{** WARNING: IEEEtranlink.bst: No hyphenation pattern has been}%
\typeout{** loaded for the language `#1'. Using the pattern for}%
\typeout{** the default language instead.}%
\else
\language=\csname l@#1\endcsname
\fi
#2}}
\providecommand{\BIBdecl}{\relax}
\BIBdecl

\bibitem{vapnik1964perceptron}
V.~Vapnik and A.~Y. Chervonenkis, ``On a perceptron class,'' \emph{Avtomatika i Telemkhanika}, vol.~25, no.~1, pp. 112--120, Feb. 1964.

\bibitem{vapnik1974theory}
V.~Vapnik and A.~Chervonenkis, ``Theory of pattern recognition,'' vol.~1, no.~1, Oct. 1974.

\bibitem{vapnik1992principles}
V.~Vapnik, ``Principles of risk minimization for learning theory,'' \emph{Advances in Neural Information Processing Systems}, vol.~4, pp. 831--838, Jan. 1992.

\bibitem{vapnik1993local}
V.~Vapnik and L.~Bottou, ``Local algorithms for pattern recognition and dependencies estimation,'' \emph{Neural Computation}, vol.~5, no.~6, pp. 893--909, Nov. 1993.

\bibitem{krzyzak1996nonparametric}
A.~Krzyzak, T.~Linder, and C.~Lugosi, ``Nonparametric estimation and classification using radial basis function nets and empirical risk minimization,'' \emph{IEEE Transactions on Neural Networks}, vol.~7, no.~2, pp. 475--487, Mar. 1996.

\bibitem{zou2009generalization}
B.~Zou, L.~Li, and Z.~Xu, ``The generalization performance of {ERM} algorithm with strongly mixing observations,'' \emph{Machine Learning}, vol.~75, no.~3, pp. 275--295, Feb. 2009.

\bibitem{esposito2021generalization}
A.~R. Esposito, M.~Gastpar, and I.~Issa, ``Generalization error bounds via {R}{\'e}nyi-, $f$-divergences and maximal leakage,'' \emph{IEEE Transactions on Information Theory}, vol.~67, no.~8, pp. 4986--5004, May 2021.

\bibitem{wang2019information}
H.~Wang, M.~Diaz, J.~C.~S. Santos~Filho, and F.~P. Calmon, ``An information-theoretic view of generalization via {W}asserstein distance,'' in \emph{Proceedings of the IEEE International Symposium on Information Theory (ISIT)}, Paris, France, Jul. 2019, pp. 577--581.

\bibitem{dalalyan2024user}
A.~S. Dalalyan and A.~G. Karagulyan, ``User-friendly guarantees for the {L}angevin {M}onte {C}arlo with inaccurate gradient,'' \emph{\normalfont{arXiv preprint arXiv:1710.00095}}, Feb. 2024.

\bibitem{li2025regularization}
W.~Li, N.~Klein, B.~Gifford, E.~Sklute, C.~Legett, and S.~Clegg, ``Regularization via $f$-divergence: An application to multi-oxide spectroscopic analysis,'' \emph{\normalfont{arXiv preprint arXiv:2502.03755}}, Feb. 2025.

\bibitem{peng2025information}
F.~Peng, M.~Zhang, and M.~Tang, ``An information-theoretic analysis for federated learning under concept drift,'' \emph{\normalfont{arXiv preprint arXiv:2506.21036}}, Jun. 2025.

\bibitem{futami2023informationtheoretic}
F.~Futami and T.~Iwata, ``Information-theoretic analysis of test data sensitivity in uncertainty,'' \emph{\normalfont{arXiv preprint arXiv:2307.12456}}, Jul. 2023.

\bibitem{huang2021stochastic}
Z.~Huang and S.~Becker, ``Stochastic gradient {L}angevin dynamics with variance reduction,'' \emph{\normalfont{arXiv preprint arXiv:2102.06759}}, Jul. 2021.

\bibitem{raginsky2016information}
M.~Raginsky, A.~Rakhlin, M.~Tsao, Y.~Wu, and A.~Xu, ``Information-theoretic analysis of stability and bias of learning algorithms,'' in \emph{Proceedings of the IEEE Information Theory Workshop (ITW)}, Cambridge, UK, Sep. 2016, pp. 26--30.

\bibitem{russo2019much}
D.~Russo and J.~Zou, ``How much does your data exploration overfit? {C}ontrolling bias via information usage,'' \emph{IEEE Transactions on Information Theory}, vol.~66, no.~1, pp. 302--323, Jan. 2019.

\bibitem{perlazaISIT2022}
S.~M. Perlaza, G.~Bisson, I.~Esnaola, A.~Jean-Marie, and S.~Rini, ``Empirical risk minimization with relative entropy regularization: {O}ptimality and sensitivity,'' in \emph{Proceedings of the IEEE International Symposium on Information Theory (ISIT)}, Espoo, Finland, Jul. 2022, pp. 684--689.

\bibitem{daunas2024TITAsymmetry}
F.~Daunas, I.~Esnaola, S.~M. Perlaza, and H.~V. Poor, ``Asymmetry of the relative entropy in the regularization of empirical risk minimization,'' \emph{IEEE Transactions on Information Theory}, vol.~71, no.~8, pp. 6198--6226, Aug. 2025.

\bibitem{perlaza2024ERMRER}
S.~M. Perlaza, G.~Bisson, I.~Esnaola, A.~Jean-Marie, and S.~Rini, ``Empirical risk minimization with relative entropy regularization,'' \emph{IEEE Transactions on Information Theory}, vol.~70, no.~7, pp. 5122 -- 5161, Jul. 2024.

\bibitem{aminian2021exact}
G.~Aminian, Y.~Bu, L.~Toni, M.~Rodrigues, and G.~Wornell, ``An exact characterization of the generalization error for the {G}ibbs algorithm,'' \emph{Advances in Neural Information Processing Systems}, vol.~34, pp. 8106--8118, Dec. 2021.

\bibitem{bu2020tightening}
Y.~Bu, S.~Zou, and V.~V. Veeravalli, ``Tightening mutual information-based bounds on generalization error,'' \emph{IEEE Journal on Selected Areas in Information Theory}, vol.~1, no.~1, pp. 121--130, Jan. 2020.

\bibitem{zou2024WorstCase}
X.~Zou, S.~M. Perlaza, I.~Esnaola, E.~Altman, and H.~V. Poor, ``The worst-case data-generating probability measure in statistical learning,'' \emph{IEEE Journal on Selected Areas in Information Theory}, vol.~5, no.~1, pp. 175 -- 189, Apr. 2024.

\bibitem{teboulle1992entropic}
M.~Teboulle, ``Entropic proximal mappings with applications to nonlinear programming,'' \emph{Mathematics of Operations Research}, vol.~17, no.~3, pp. 670--690, Aug. 1992.

\bibitem{beck2003mirror}
A.~Beck and M.~Teboulle, ``Mirror descent and nonlinear projected subgradient methods for convex optimization,'' \emph{Operations Research Letters}, vol.~31, no.~3, pp. 167--175, Jan. 2003.

\bibitem{alquier2021non}
P.~Alquier, ``Non-exponentially weighted aggregation: Regret bounds for unbounded loss functions,'' in \emph{Proceedings of the 38th International Conference on Machine Learning (ICML)}, vol. 139, Jul. 2021, pp. 207--218.

\bibitem{perlazaISIT2024a}
F.~Daunas, I.~Esnaola, S.~M. Perlaza, and H.~V. Poor, ``Equivalence of empirical risk minimization to regularization on the family of $f$-divergences,'' in \emph{Proceedings of the IEEE International Symposium on Information Theory (ISIT)}, Athens, Greece, Jul. 2024.

\bibitem{wei2021optimizing}
J.~Wei and Y.~Liu, ``When optimizing $f$-divergence is robust with label noise,'' \emph{\normalfont{arXiv preprint arXiv:2011.03687}}, Aug. 2021.

\bibitem{liu2023smoothed}
Z.~Liu, B.~P. Van~Parys, and H.~Lam, ``Smoothed $f$-divergence distributionally robust optimization,'' \emph{\normalfont{arXiv preprint arXiv:2306.14041}}, Oct. 2023.

\bibitem{namkoong2016stochastic}
H.~Namkoong and J.~C. Duchi, ``Stochastic gradient methods for distributionally robust optimization with $f$-divergences,'' \emph{Advances in Neural Information Processing Systems}, vol.~26, pp. 1476--1485, Dec. 2016.

\bibitem{hu2013kulback}
Z.~Hu and L.~J. Hong, ``{K}ullback-{L}eibler divergence constrained distributionally robust optimization,'' \emph{Available at Optimization Online}, vol.~1, no.~2, Nov. 2013.

\bibitem{hastings1970monte}
W.~K. Hastings, ``{M}onte {C}arlo sampling methods using {M}arkov chains and their applications,'' \emph{Biometrika}, vol.~57, no.~1, pp. 97--109, May 1970.

\bibitem{gilks1992adaptive}
W.~R. Gilks and P.~Wild, ``Adaptive rejection sampling for {G}ibbs sampling,'' \emph{Journal of the Royal Statistical Society: Series C (Applied Statistics)}, vol.~41, no.~2, pp. 337--348, Jun. 1992.

\bibitem{perlazaISIT2023b}
S.~M. Perlaza, I.~Esnaola, G.~Bisson, and H.~V. Poor, ``On the validation of {G}ibbs algorithms: {T}raining datasets, test datasets and their aggregation,'' in \emph{Proceedings of the IEEE International Symposium on Information Theory (ISIT)}, Taipei, Taiwan, Jun. 2023.

\bibitem{bulatov2005complexity}
A.~Bulatov and M.~Grohe, ``The complexity of partition functions,'' \emph{Theoretical Computer Science}, vol. 348, no.~2, pp. 148--186, Sep. 2005.

\bibitem{bulatov2013complexity}
A.~A. Bulatov, ``The complexity of the counting constraint satisfaction problem,'' \emph{Journal of the {ACM (JACM)}}, vol.~60, no.~5, pp. 1--41, Sep. 2014.

\bibitem{mcquillan2013computational}
C.~McQuillan, ``The computational complexity of approximation of partition functions,'' PhD thesis, University of Liverpool, Liverpool, UK, Jun. 2013.

\bibitem{rockafellar1970conjugate}
R.~T. Rockafellar, \emph{Conjugate Convex Functions in Optimal Control and the Calculus of Variations}, 2nd~ed.\hskip 1em plus 0.5em minus 0.4em\relax Princeton, NJ, USA: Princeton University Press, 1970.

\bibitem{boyd2004convex}
S.~Boyd, S.~P. Boyd, and L.~Vandenberghe, \emph{Convex optimization}, 1st~ed.\hskip 1em plus 0.5em minus 0.4em\relax Cambridge, UK: Cambridge University Press, 2004.

\bibitem{oswaldo2013TIFT}
O.~de~Oliveira, ``{The Implicit and Inverse Function Theorems: Easy Proofs},'' \emph{Real Analysis Exchange}, vol.~39, no.~1, pp. 207 -- 218, 2013.

\bibitem{csiszar1967information}
I.~Csisz{\'a}r, ``Information-type measures of difference of probability distributions and indirect observation,'' \emph{Studia Scientiarum Mathematicarum Hungarica}, vol.~2, no.~1, pp. 299--318, Jun. 1967.

\bibitem{luenberger1964observing}
D.~G. Luenberger, ``Observing the state of a linear system,'' \emph{IEEE Transactions on Military Electronics}, vol.~8, no.~2, pp. 74--80, Apr. 1964.

\bibitem{luenberger1997bookOptimization}
------, \emph{Optimization by Vector Space Methods}, 1st~ed.\hskip 1em plus 0.5em minus 0.4em\relax New York, NY, USA: Wiley, 1997.

\bibitem{bartle2000introduction}
R.~G. Bartle and D.~R. Sherbert, \emph{Introduction to Real Analysis}, 3rd~ed.\hskip 1em plus 0.5em minus 0.4em\relax New York, NY, USA: Wiley New York, 2000.

\bibitem{douchet2010Analyse}
J.~Douchet, \emph{Analyse : {R}ecueil d'Exercices et Aide-M\'emoire}, 3rd~ed.\hskip 1em plus 0.5em minus 0.4em\relax Lausanne, Switzerland: PPUR, 2010, vol.~1.

\bibitem{InriaRR9594}
F.~Daunas, I.~Esnaola, and S.~M. Perlaza, ``A dual optimization view to empirical risk minimization with $f$-divergence regularization,'' INRIA, Centre Inria d'Université Côte d'Azur, Sophia Antipolis, France, Tech. Rep. RR-9594, Jul. 2025.

\bibitem{ash2000probability}
R.~B. Ash and C.~A. Doleans-Dade, \emph{Probability and Measure Theory}, 2nd~ed.\hskip 1em plus 0.5em minus 0.4em\relax Burlington, MA, USA: Academic Press, 2000.

\end{thebibliography}

\end{document}